\documentclass[sigconf,nonacm, authorversion]{acmart}

\usepackage{natbib}
\setcitestyle{numbers,square}

\usepackage{array}
\usepackage{booktabs} 
\usepackage{multirow} 
\usepackage{graphicx} 
\usepackage{float}
\usepackage{tabularx}
\usepackage{pifont}   
\usepackage{xcolor}   
\usepackage{colortbl}
\usepackage{geometry}
\usepackage{xcolor}
\usepackage{soul}     
\newcommand{\cmark}{\ding{51}}%
\newcommand{\xmark}{\ding{55}}%

\usepackage[most]{tcolorbox} 
\usepackage{listings}
\usepackage{xcolor}
\usepackage{caption} 

\lstdefinestyle{mypromptstyle}{
    basicstyle=\ttfamily\scriptsize,
    breaklines=true,
    columns=fullflexible,
    frame=none,
    backgroundcolor=\color{white},
    moredelim=[is][\bfseries]{**}{**}, 
    moredelim=[is][\color{blue}]{```json}{```}, 
}

\usepackage{hyperref}
\begin{document}

\title{CDRRM: Contrast-Driven Rubric Generation for Reliable and Interpretable Reward Modeling}

\author{Dengcan Liu}
\authornote{Both authors contributed equally to this research.}
\email{ldc123@mail.ustc.edu.cn}
\affiliation{%
  \institution{University of Science and Technology of China}
  \city{Hefei}
  \country{China}
}

\author{Fengkai Yang}
\authornotemark[1]
\email{yangfengkai@stu.pku.edu.cn}
\affiliation{%
  \institution{Peking University}
  \city{Beijing}
  \country{China}
}

\author{Xiaohan Wang$^\dagger$}
\affiliation{%
 \institution{Meituan}
 \city{Beijing}
 \country{China}
}

\author{Shurui Yan}
\affiliation{%
  \institution{Meituan}
  \city{Beijing}
  \country{China}
}

\author{Jiajun Chai}
\affiliation{%
  \institution{Meituan}
  \city{Beijing}
  \country{China}
}

\author{Jiahao Li}
\affiliation{%
  \institution{University of Science and Technology of China}
  \city{Hefei}
  \country{China}
}

\author{Yikun Ban}
\affiliation{%
  \institution{BeiHang University}
  \city{Beijing}
  \country{China}
}

\author{Zhendong Mao$^\dagger$}
\affiliation{%
  \institution{University of Science and Technology of China}
  \city{Hefei}
  \country{China}
}

\author{Wei Lin}
\affiliation{%
  \institution{Meituan}
  \city{Beijing}
  \country{China}
}

\author{Guojun Yin}
\affiliation{%
  \institution{Meituan}
  \city{Beijing}
  \country{China}
}


\begin{abstract}
Reward modeling is essential for aligning Large Language Models(LLMs) with human preferences, yet conventional reward models suffer from poor interpretability and heavy reliance on costly expert annotations. While recent rubric-based approaches enhance evaluation transparency, they lack systematic quality control, yielding noisy and redundant criteria, failing to mitigate persistent biases (e.g., verbosity, position) in LLM evaluators, and creating a scalability-reliability trade-off. To address these limitations, we propose \textbf{CDRRM} (\textbf{C}ontrast-\textbf{D}riven \textbf{R}ubric \textbf{R}eward \textbf{M}odel), a framework built on a novel \textbf{Contrast-then-Synthesis} paradigm for high-quality rubric generation and guided preference judgment. CDRRM first conducts multi-dimensional contrastive profiling on preference pairs to identify causal discriminative factors, then synthesizes these insights into compact, context-aware rubrics to guide preference judgments. Extensive experiments on three authoritative benchmarks (RewardBench, RMBench, RMB) demonstrate that CDRRM achieves state-of-the-art performance across diverse domains and effectively mitigates aforementioned evaluation biases. Notably, our approach delivers exceptional data efficiency: training the rubric generator on only 3k high-quality samples empowers a frozen pre-trained judge model to outperform fully fine-tuned baselines. This work offers a scalable, interpretable, and data-efficient path for reward modeling.
\end{abstract}




\maketitle

\footnotetext{$^\dagger$ Corresponding author}
\footnotetext[2]{$^\ddagger$ Code is available at: https://github.com/ldcan/CDRRM.git}

\section{Introduction}
Reward modeling is a cornerstone of post-training for aligning large language models (LLMs) with human preferences \cite{DBLP:journals/corr/abs-2405-16455, DBLP:journals/corr/abs-2401-06080, DBLP:journals/corr/abs-2601-08521}. While traditional scalar reward models have long served as a classic technical solution for early LLM alignment tasks \cite{christiano2017deep}, they suffer from two critical limitations that undermine their utility for advanced alignment scenarios. First, their inherent opacity leads to a "black box" evaluation process with no explicit rationale for preference decisions, exposing such models to the risk of \textit{reward hacking} \cite{DBLP:conf/iclr/PanBS22, DBLP:conf/icml/ReberRNGV25}. Second, training robust scalar models relies heavily on large-scale high-quality expert annotations \cite{ouyang2022training,bai2022training}, which imposes severe scalability and domain adaptability bottlenecks for large-scale alignment deployments. To address these limitations—and to meet the growing demand for interpretable, transparent evaluation in the emerging LLM-as-a-Judge paradigm—research community has witnessed a rapid shift toward Generative Reward Models (GenRMs) \cite{zhu2023judgelm,deepseek-grm, rm-r1}. GenRMs generate explicit reasoning traces, structured critiques, and judgment justifications to ground their preference decisions, thereby drastically enhancing the transparency and interpretability of LLM-as-a-Judge evaluations.

Within this generative paradigm, rubric-based reward modeling has attracted considerable attention as a principled approach\cite{openrubrics, rm-r1, r3, autorubric, DBLP:journals/corr/abs-2511-07896}. By decomposing complex judgments into structured, semantic rubrics, these methods offer greater transparency and precision in evaluation. However, constructing high-quality rubrics remains a core bottleneck. Current approaches largely rely on either labor-intensive manual annotation \cite{DBLP:journals/corr/abs-2511-10507} or direct prompting of LLMs \cite{openrubrics, DBLP:journals/corr/abs-2509-15110}, both of which have notable shortcomings: manual curation is not scalable, while direct prompting often yields noisy, redundant rubrics weakly related to actual discriminative factors. Furthermore, existing methods have not effectively addressed persistent biases inherent
in LLM evaluators(e.g., verbosity bias, stylistic preference bias, and position bias)\cite{verbosity,autorubric}, which continue to erode the reliability of the
alignment process.

\begin{figure*}[t]
  \centering
  \includegraphics[width= \linewidth,trim=0 70 0 100,clip]{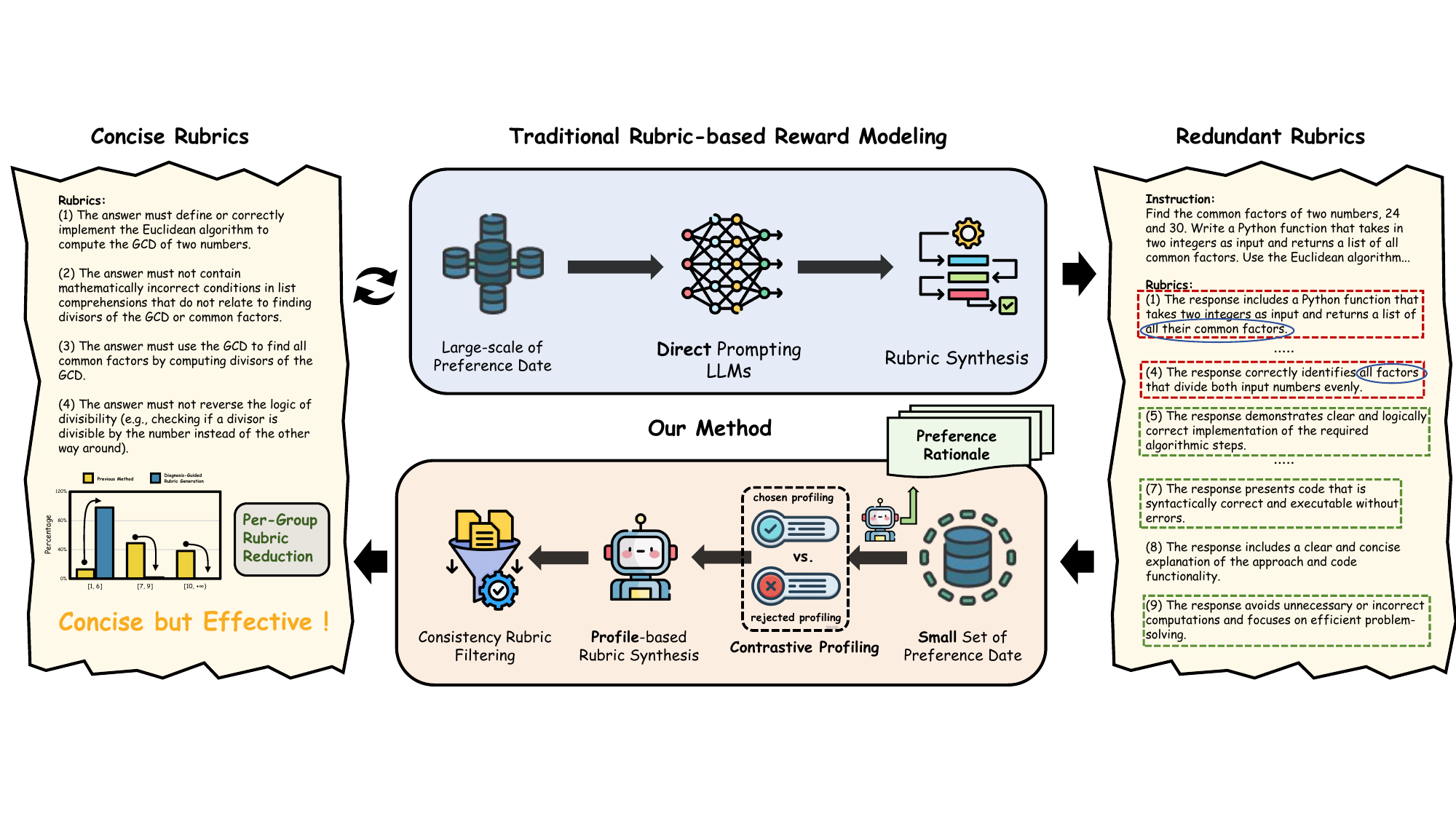}
  \caption{An illustrative example of rubric generation for a Greatest Common Divisor (GCD) task, contrasting rubrics from direct prompting (right, redundant and potentially misleading) with those from our Contrast-then-Synthesis paradigm (left, concise and effective). The bottom-left panel shows statistics on the number of generated rubrics with respect to single preferences.}
  \Description{A short description for accessibility.}
  \label{fig:intro}
\end{figure*}

In this work, we aim to endow rubric generation models with the capability to yield succinct and highly efficacious rubrics, enabling more robust guidance for reward modeling and effective mitigation of inherent biases in LLM evaluators. To this end, we introduce a structured framework comprising \textbf{Contrastive Profiling} and \textbf{Rubric Synthesis}. This paradigm moves beyond generic rubric generation by conducting rigorous contrastive analysis of preference pairs to pinpoint the exact causal factors for preference decisions, aiming to pinpoint exactly why a response is chosen or rejected. Specifically, we employ an LLM-as-a-Judge to perform a task-aligned multi-dimensional contrastive analysis, explicitly isolating evidence-based causal factors(e.g., factual errors, logical gaps) that drive preference judgments. These differential insights are then synthesized into concise, high-impact rubrics, filtering out noise and redundancy inherent in raw model outputs. Building on this high-fidelity rubric dataset, we propose the \textbf{C}ontrast-\textbf{D}riven \textbf{R}ubric \textbf{R}eward \textbf{M}odel (\textbf{CDRRM}), which instantiates this paradigm through two specialized, mutually coupled components: a \textbf{Rubric Generator}, trained to synthesize context-aware evaluation criteria, and a \textbf{Judge Model}, fine-tuned to predict preferences strictly conditioned on these rubrics.

We conduct extensive evaluations on three authoritative benchmarks: RewardBench \cite{rewardbench}, RMBench \cite{rmbench}, and RMB \cite{rmb}. Empirical results demonstrate that CDRRM achieves state-of-the-art performance across diverse domains and significantly mitigates persistent evaluation biases such as verbosity and position biases. Most notably, our method exhibits exceptional data efficiency: training the Rubric Generator on just 3k high-quality samples enables a frozen base model—guided solely by these synthesized rubrics—to outperform fully fine-tuned baselines.

Overall, our main contributions are as follows:
\begin{itemize}
    \item We propose \textbf{Contrast-then-Synthesis}, a novel paradigm that transforms opaque preference modeling into an explicit, rubric-guided reasoning process. By grounding rubric generation in rigorous contrastive profiling of preference pairs, our method systematically isolates task-critical discriminative factors, eliminating redundant evaluation criteria and mitigating the hallucination of irrelevant assessment standards at its root.
    \item We introduce \textbf{CDRRM}, a concrete instantiation of the Contrast-then-Synthesis paradigm that synthesizes precise, concise rubrics to guide preference judgments. It enables robust, interpretable and generalizable preference evaluation across diverse domains, and we will release our two-stage dataset publicly to support future research.
    \item We conduct extensive evaluations across three benchmarks, demonstrating that CDRRM establishes a new state-of-the-art in reward modeling. Compared to rubric-based baselines, CDRRM improves average accuracy by 5.7\% across all benchmarks and achieves a remarkable 18\% gain on RMBench Hard. 
\end{itemize}

\section{Preliminaries}
\subsection{Rubric Learning}
In this paper, we adopt pairwise setting on reward modeling \cite{DBLP:conf/iclr/0017ST25}, given the preference dataset $\mathcal{D} = \left \{ x_{i}, y_{i}^{c} , y_{i}^{r}\right \}_{i=1}^{N}$, where $x$ denotes an input prompt and $(y^c, y^r)$ is a response pair consisting of the chosen and rejected responses, respectively. The pair-wise training paradigm is built on the Bradley-Terry model \cite{bradley1952rank}, the formulation can be modeled as:
\begin{equation}
    \mathbb{P}\!\left(y^c \succ y^r \mid x\right)
= \sigma\!\left(r_{\theta}(x, y^c) - r_{\theta}(x, y^r)\right)
\end{equation}
Its objective is to optimize the opaque, black-box reward function $r_{\theta}$ for reward model training—a paradigm that inherently predisposes the model to reward hacking. Rubric learning builds a structured framework of evaluation criteria customized for the given prompt $x$. We formalize the collection of criteria spanning various dimensions as:
\begin{equation}
    \mathcal{R}\left(x\right) = \{ r_1, r_2, \ldots, r_k \}=\{r_i\}_{i=1}^k
\end{equation}
where each criteria $r_i$ represents an individual rubric item, with each description precisely delineating a targeted dimension of response quality to be evaluated. And the reward function can be defined as:
\begin{equation}
    R\left(x, y^{c} , y^{r}\right)=r_{\theta}\left(x,y^{c} , y^{r},\{r_i\}_{i=1}^k\right) 
\end{equation}
The rubric-based reward integrates criteria across multiple dimensions, yielding a transparent and interpretable evaluation. The quality of generated rubrics is crucial for the rubric learning. However, prior rubric-generation methods are often overly coarse-grained, resulting in redundant and overlapping rubrics. This issue is summarized as follows.

\subsection{Problem Statement}
Under the pairwise evaluation paradigm, previous rubric-based methods rely solely on direct prompting to elicit rubrics. However, by attempting to generate criteria in a single step without prior fine-grained analysis, the model lacks intrinsic alignment with discriminative human standards. This limitation introduces non-trivial redundancy and spurious noise: the resulting rubrics are plagued by overlapping semantics and irrelevant details, flaws that can significantly misguide the reward model's training.

In this section, we systematically analyze the redundancy inherent in traditional rubric generation methods\cite{openrubrics}. Empirical evidence from existing rubric-based datasets supports our hypothesis: as shown in Figure \ref{fig:intro}, the majority of samples contain at least seven rubrics. This excessive quantity contradicts the finding that preference judgments typically hinge on a sparse set of salient factors\cite{gigerenzer2002bounded}. To quantify this redundancy, we conducted a perturbation study by randomly masking one to three rubrics (along with their rationales) from the training data and retraining the Reward Model to observe the impact on performance.

Table \ref{tab:remove} reveals that aggressively pruning these rubrics results in negligible performance degradation, with a maximum deviation of only 0.42\% on the validation set. This finding confirms that a significant portion of the generated rubrics in current datasets creates non-trivial redundancy rather than providing informative signals. These observations underscore the necessity of a more selective generation process, motivating our proposed Contrast-then-Synthesis strategy, which we detail in the subsequent section.

\begin{figure*}[t]
  \centering
  \includegraphics[width= \linewidth,trim=180 110 90 0,clip]{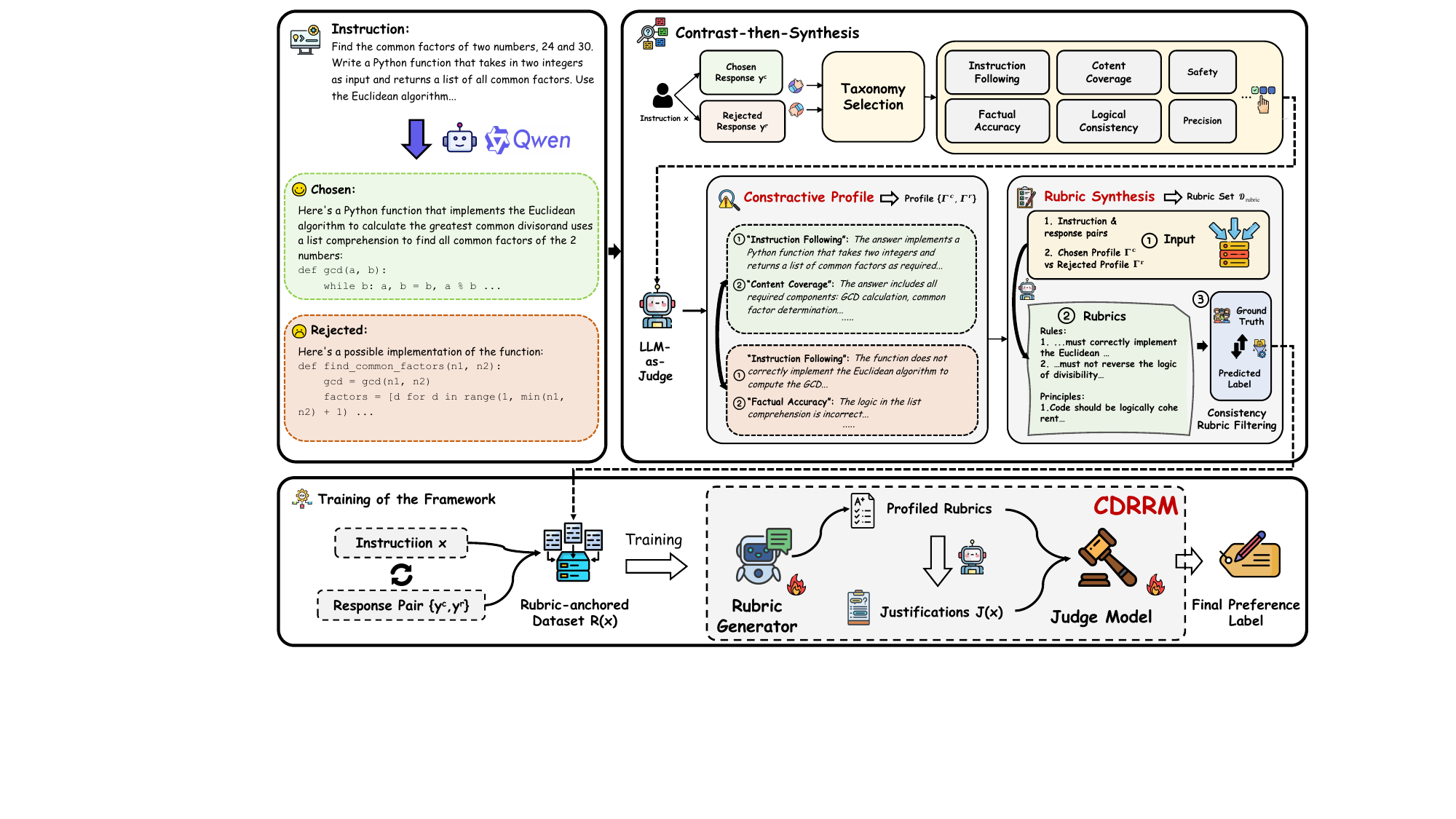}
  \caption{The CDRRM framework. (Top) The Contrast-then-Synthesis paradigm synthesizes evidence-based rubrics via contrastive analysis of preference pairs. (Bottom) These rubrics, paired with synthesized rubric-grounded justifications, supervise the training of a Rubric Generator (to automate context-aware criterion synthesis) and a Judge Model (to generate rubric-aligned justifications for precise preference predictions).}
  \label{fig:main-framework}
\end{figure*}

\section{Methodology}
In this section, we introduce \textbf{Contrast-then-Synthesis}, a novel framework tailored for generating high-quality rubrics to guide reward modeling. As illustrated in Figure \ref{fig:main-framework}, our approach contains two synergistic components: (1) Contrastive Profiling, which conducts multi-dimensional contrastive analysis on preference pairs to isolate core discriminative factors that drive judgments; (2) Rubric Synthesis, which summarizes these diagnostic insights into concise evaluation criteria. Building on these two steps, we train a dedicated rubric generator that produces precise, non-redundant rubrics to facilitate reliable preference discrimination.

\begin{table}[t]
  \caption{Impact of Rubric Reduction on Accuracy.}
  \label{tab:remove}
  \centering
  {\setlength{\tabcolsep}{10pt} 
  \begin{tabular}{lcccc}
    \toprule
    Del & \textbf{Baseline} & \textbf{Del 0} & \textbf{Del 1} & \textbf{Del 2} \\
    \midrule
    Acc.\ (\%)    & 95.65 & 95.70 & 95.23 & 95.37 \\
    $\Delta$ (\%) & 0.00  & +0.05 & -0.42 & -0.28 \\
    \bottomrule
  \end{tabular}
  }
  \Description{A short description of the table for accessibility.}
\end{table}

\subsection{Contrastive Profiling}

\par \textbf{Adaptive Evaluation Taxonomy.} We initially establish a rigorous taxonomy set 
$\mathcal{T}$ to comprehensively dissect each response's quality. Rather than adopting a static set of evaluative criteria, we employ a dynamic evaluation strategy that selectively activates only the dimensions germane to the specific instruction context. Let $\mathcal{T} = \{d_1, \dots, d_m\}$ denote the full spectrum of analysis dimensions (e.g., Instruction Following, Logical Consistency, Safety, $\dots$). Given an instruction 
$x$ and a response $y$, the model first identifies the active dimension subset $\mathcal{T}_{x,y} \subseteq \mathcal{T}$:

\begin{equation}
\mathcal{T}_{x,y} = \text{Select}(\mathcal{T}=\{d_1, \dots, d_m\} \mid x,y)
\end{equation}

\noindent This dynamic selection mechanism ensures the analysis focuses on salient quality factors, effectively minimizing noise in the subsequent synthesis process. With the selected taxonomy, the LLM-as-a-Judge generates analysis $\gamma_{x,y}$ across these dimensions:
\begin{equation}
    \gamma_t = \mathrm{Judge}(x, y, d_t), \quad \forall d_t \in \mathcal{T}_{x,y}
\end{equation}

\noindent \textbf{Evidence-Anchored Verification.}
To further ensure the verifiability of the analysis, while precluding ambiguity in evaluative criteria and the hallucination \cite{DBLP:journals/corr/abs-2412-05579}, we enforce the \textbf{Evidence-Anchored Constraint}. Instead of generating abstract assessments, for each active dimension $d_t \in \mathcal{T}_{x,y}$, the model is mandated to ground its judgments in original text spans. And we have the evidence-anchored analysis:

\begin{equation}
\gamma_t' \triangleq \bigl(\gamma_t,\ \hat{s}^{x}_t,\ \hat{s}^{y}_t\bigr)
\end{equation}

where $\hat{s}^{x}_t$ is the specific constraint in the instruction (e.g., "no python code") and $\hat{s}^{y}_t$ is the corresponding segment in the response (e.g., a python code block). Aggregating these evidence triplets, we formalize the profile for each instance and the resulting dataset as:
\begin{equation}
    \Gamma \triangleq \{\, (d_t,\gamma_t') \mid d_t \in \mathcal{T}_{x,y} \,\},  ~~\mathcal{D}_{\mathcal{T}} = \left\{ x_i,\ y_i^c,\ y_i^r,\ \Gamma_{i}^{c},\ \Gamma_{i}^{r} \right\}_{i=1}^{N}
\end{equation}

This structured profiling strategy mandates that subsequent rubric generation be anchored to factual observations rather than model priors, thereby significantly enhancing the interpretability and discriminability of the synthesized rubrics.

\subsection{Rubric Synthesis}

By decomposing preference judgments into factual, multi-dimensional evidence, our profiles provide a transparent, fact-based basis for evaluation. In contrast to direct prompting approaches that rely on implicit inference, \textbf{Rubric Synthesis} leverages these explicit, differential insights to generate rubrics directly grounded in the observed quality gaps between preference pairs. Specifically, we formulate this process as a conditional generation task, where a dedicated teacher LLM generates a concise rubric set $\mathcal{R}(x_i)$ that best explains the discrepancy between the chosen profile $\Gamma_i^c$ and the rejected profile $\Gamma_i^r$. Formally, this generation process is expressed as maximizing the likelihood of the rubric set given the instruction and contrastive profiles:
\begin{equation}
\mathcal{R}(x_i) = \arg\max_{\mathcal{R}} P_{\text{Teacher-LLM}}\left(\mathcal{R} \mid x_i, \Delta\left(\Gamma_i^c, \Gamma_i^r\right)\right)
\end{equation}
where $\Delta(\Gamma_i^c, \Gamma_i^r)$ denotes the structured contrastive concatenation operator that splices the chosen profile $\Gamma_i^c$ and rejected profile $\Gamma_i^r$ into a unified text sequence, thus explicitly highlighting the discriminative factors underlying human preference judgments. This design ensures that each generated rubric is tightly aligned with the core rationale of human preference judgments, thereby effectively eliminating redundant rubrics induced by the model’s inherent priors. \\

\noindent \textbf{Consistency Filtering and Dataset Construction.} To further guarantee the robustness of the generated rubric sets and eliminate extraneous noisy rubrics, we enforce a \textbf{Preference-Consistency Constraint}. Specifically, we prompt the LLM to re-evaluate the preference pair $\{y_i^c, y_i^r\}$ under the strict condition of the generated rubric $\mathcal{R}(x_i)$, and only retain rubric sets for which the predicted preference label $\hat{l}_i$ matches the ground-truth preference label $l_i$. We formalize this validity indicator as:
\begin{equation}
    \mathbb{I}_{\text{valid}}(\mathcal{R}(x_i)) = 
    \begin{cases} 
    1 & \text{if } \text{Judge}\left(x_i, y_i^c, y_i^r \mid \mathcal{R}(x_i)\right) = l_i \\
    0 & \text{otherwise}
    \end{cases}
\end{equation}
After filtering out rubric sets that fail this consistency check, we construct a high-quality, insight-based rubric dataset $\mathcal{D}_{\text{rubric}}$, which serves as supervised data for training our Rubric Generator. Formally, the dataset is defined as:
\begin{equation}
    \mathcal{D}_{\text{rubric}} = \left\{ (x_i, y_i^c, y_i^r, \mathcal{R}(x_i)) \mid \mathbb{I}_{\text{valid}}(\mathcal{R}(x_i)) = 1 \right\}_{i=1}^{N}
\end{equation}

\begin{table*}[ht]
\caption{Main results on RewardBench, RMBench, and RMB. We report the accuracy scores for each category. The best results in each column are highlighted in \textbf{bold}. ``Help'' and ``Harm'' denote Helpfulness and Harmlessness, respectively. The ``Average'' column represents the mean of the overall scores across the three benchmarks.}
\footnotesize
\centering
\resizebox{\textwidth}{!}{
\begin{tabular}{l|c|cccc|ccc|c}
\toprule
\multirow{2}{*}{\textbf{Models}} & \textbf{RewardBench} & \multicolumn{4}{c|}{\textbf{RMBench}} & \multicolumn{3}{c|}{\textbf{RMB}} & \multirow{2}{*}{\textbf{Avg.}} \\
\cmidrule(lr){2-2} \cmidrule(lr){3-6} \cmidrule(lr){7-9} 
 & Overall & Easy & Medium & Hard & Overall & Help & Harm & Overall & \\
\midrule
\multicolumn{10}{l}{\textit{\textbf{Scalar RMs}}} \\
SteerLM-RM-70B & 88.8 & 48.3 & 54.9 & 54.3 & 52.5 & 57.4 & 67.3 & 62.4 & 67.9 \\
InternLM2-20B-Reward & 90.2 & 82.6 & 71.6 & 50.7 & 68.3 & 76.3 & 67.0 & 71.7 & 76.7 \\
ArmoRM-Llama3-8B-v0.1 & 90.4 & 82.2 & 71.0 & 49.8 & 67.7 & 78.7 & 66.3 & 72.5 & 76.9 \\
Skywork-Reward-Llama-3.1-8B & 92.5 & 89.0 & 74.7 & 46.6 & 70.1 & 78.1 & 75.9 & 77.0 & 79.9 \\
INF-ORM-Llama3.1-70B & \textbf{95.1} & \textbf{91.8} & 76.1 & 44.8 & 70.9 & 79.8 & 76.7 & 78.3 & 81.4 \\
\midrule
\multicolumn{10}{l}{\textit{\textbf{GenRMs}}} \\
BR-RM-Qwen-8B & 91.0 & 91.7 & 87.3 & 76.1 & 85.0 & 76.9 & \textbf{82.2} & 79.6 & 85.2 \\
DeepSeek-GRM-27B & 86.0 & 84.6 & 76.5 & 57.0 & 72.7 & 80.5 & 76.1 & 78.3 & 79.0 \\
Skywork-Critic-Llama-3.1-70B & 93.3 & 85.6 & 73.7 & 56.5 & 71.9 & 75.3 & 61.4 & 68.4 & 77.9 \\
\midrule
\multicolumn{10}{l}{\textit{\textbf{Rubric-based RMs}}} \\
RUBRIC-RM-8B & - & - & - & - & 62.2 & - & - & - & - \\
RM-R1-Qwen-Instruct-32B & 91.4 & 86.3 & 80.5 & 70.4 & 79.1 & 79.1 & 80.9 & 80.0 & 83.5 \\
R3-Qwen3-8B & 88.8 & 89.0 & 83.4 & 71.9 & 81.4 & - & - & - & - \\
\midrule
\multicolumn{10}{l}{\textit{\textbf{Ours (CDRRM)}}} \\
\rowcolor{green!5}
CDRRM-8B (Base) & 90.4 & 90.4 & 86.8 & 81.1 & 86.1 & 85.3 & 76.3 & 80.8 & 85.8 \\
\rowcolor{blue!5}
CDRRM-8B (SFT) & 92.0 & 90.0 & 86.3 & 81.0 & 85.8 & 88.1 & 78.4 & 83.2 & 87.0 \\
\rowcolor{green!5}
CDRRM-14B (Base) & 92.5 & 89.9 & 87.4 & 82.5 & 86.6 & 86.5 & 79.7 & 83.7 & 87.6 \\
\rowcolor{blue!5}
CDRRM-14B (SFT) & 92.8 & 90.9 & \textbf{88.6} & \textbf{83.4} & \textbf{87.6} & \textbf{88.6} & 80.3 & \textbf{84.4} & \textbf{88.3} \\
\bottomrule
\end{tabular}%
}
\label{tab:main_results}
\end{table*}

\subsection{Model Training}
With the high-fidelity dataset $\mathcal{D}_{rubric}$ constructed via the Contrast-then-Synthesis paradigm,  we move to the training phase of our framework, which involves two specialized components: a \textbf{Rubric Generator} for automated criteria synthesis, and a \textbf{Judge Model} for rubric-guided preference evaluation.\\

\noindent \textbf{Rubric Generator Training.} To distill the teacher model's rubric generation capabilities into a more efficient student Rubric Generator, we train the latter on the validated dataset $\mathcal{D}_{\text{rubric}}$. Conditioned on the input $(x_i, y_i^c, y_i^r)$, the model autoregressively predicts the corresponding $\mathcal{R}(x_i)$, with the training objective defined as minimizing the negative log-likelihood:

\begin{equation}
    \mathcal{L}_{\text{gen}}(\phi)
    =
    -\mathbb{E}_{\mathcal{D}_{\text{rubric}}}
    \left[
    \sum_{t=1}^{|\mathcal{R}|}
    \log q_{\phi}\!\left(\mathcal{R}_t \mid x, y_i^c, y_i^r, \; \mathcal{R}_{<t}\right)
    \right]
\end{equation}

where $\phi$ denotes the parameters of the Rubric Generator. This loss trains the generator to produce fine-grained, context-aware rubrics that convert implicit evaluation requirements into explicit criteria for subsequent preference judgment. \\

\noindent \textbf{Judge Model Training.} To enhance the discriminative power of rubrics for preference ranking, we build on the pre-trained Rubric Generator to construct a specialized dataset for the Judge Model and fine-tune the model accordingly. Specifically, we first leverage the teacher model to generate justifications for preference pairs, which are conditioned on the rubrics generated by the Rubric Generator, this process is formally denoted as:

\begin{equation}
    \mathcal{J}(x_i) = \text{Judge}\left(x_i,\ y_i^c,\ y_i^r \mid \mathcal{R}(x_i)\right)
\end{equation}

We integrate these rubric-guided justifications into the original rubric dataset $\mathcal{D}_{\text{rubric}}$ to construct the training dataset for the Judge Model:
\begin{equation}
    \mathcal{D}_{\text{judge}} = \{ (x_i, y_i^c, y_i^r, \mathcal{R}(x_i), \mathcal{J}(x_i)) \}_{i=1}^{N}
\end{equation}

We then fine-tune the Judge Model (parameterized by $\theta$) to first generate such justifications autoregressively before making the final preference decision, ensuring its judgments are explicitly grounded in the provided rubrics. The corresponding training objective is formulated as minimizing the negative log-likelihood:
\begin{equation}
    \mathcal{L}_{\text{judge}}(\theta)
    =
    -\mathbb{E}_{\mathcal{D}_{\text{judge}}}
    \left[
    \sum_{t=1}^{|\mathcal{J}|}
    \log q_{\theta}\!\left(\mathcal{J}_{t}
    \,\middle|\,
    x, y^c, y^r, \mathcal{R}, \mathcal{J}_{<t}
    \right)
    \right]
\end{equation}

\section{Experiments}

\subsection{Datasets and Experiment Settings}
\noindent\textbf{Data Sources.} 
We leverage the OpenRubrics dataset \cite{openrubrics} as the foundation for our experiments. This comprehensive corpus comprises 35.6k samples derived from a diverse integration of public preference and instruction-tuning datasets. It spans both general conversational domains—sourced from UltraFeedback \cite{DBLP:conf/icml/CuiY0YH0NXXL0024}, Tulu 2.5 \cite{DBLP:conf/nips/IvisonW0WP0S0H24}, and HelpSteer3 \cite{DBLP:journals/corr/abs-2505-11475}—and specialized scientific fields, including physics and medicine from MegaScience \cite{DBLP:journals/corr/abs-2507-16812} and diagnostic reasoning from Medical01 \cite{DBLP:journals/corr/abs-2412-18925}. This broad coverage ensures that our models are trained on a robust distribution of instruction types.\\

\noindent\textbf{Training Data Construction.} 
We construct our training data in two distinct phases, adhering to the Contrast-then-Synthesis paradigm.
\begin{itemize}
    \item \textbf{Rubric Generator Data:} We first sample a subset of 3,000 instructions with their corresponding response pairs. Utilizing \textbf{Qwen3-235B-A22B-Instruct} \cite{qwen3} as the teacher model, we synthesize high-fidelity discriminative rubrics via the Contrastive Profiling process described in Section 3. These rubrics serve as the ground-truth targets for training the Rubric Generator.
    \item \textbf{Judge Model Data:} To train the judge model, we sample another subset of 3,000 instances. For each instance, we employ the trained Rubric Generator to produce instruction-specific rubrics. We then prompt the teacher model (Qwen3-235B) to generate detailed justifications and preference labels conditioned on these rubrics, creating a rubric-grounded judgment dataset.
\end{itemize}
The impact of data scaling on model performance is systematically analyzed in Section 5.3.\\

\noindent\textbf{Model Backbones.} 
In our primary experiments, we employ Qwen3-8B as the foundational backbone for both the Rubric Generator and the Judge Model. To explore the scalability of our framework, we further extend our evaluation to larger model, specifically Qwen3-14B. Unless explicitly stated otherwise, all results reported in subsequent sections are derived from the Qwen3-8B variant. We use Swift\cite{ms-swift} for training CDRRM (via SFT). For evaluation, we use the benchmarks’ official scripts
where available. To facilitate reproducibility, we release our training and inference configuration in
Appendix \ref{sec:app_para}. Prompts, including rubric templates, are provided in Appendix \ref{sec:prompts}.\\

\noindent\textbf{Baselines and Evaluation Benchmarks.} 
We evaluate CDRRM via comprehensive benchmarking against a broad range of state-of-the-art reward models, clustered into three well-defined paradigmatic categories::
\begin{itemize}
    \item \textbf{Scalar RMs:} Representing traditional score-based approaches, we select top-performing models including SteerLM-RM-70B \cite{steerlm}, InternLM2-20B-Reward \cite{internlm}, Skywork-Reward-Llama-3.1-8B
    \cite{skywork}, INF-ORM-Llama3.1-70B, and ArmoRM-Llama3-8B-v0.1 \cite{armorm}.
    \item \textbf{Generative RMs (GenRMs):} We compare against models that output natural language critiques or reasoning traces, specifically Skywork-Critic-Llama-3.1-70B \cite{skywork}, BR-RM \cite{br-rm}, and DeepSeek-GRM-27B-RFT \cite{deepseek-grm}.
    \item \textbf{Rubric-based RMs:} To highlight the advantages of our Contrast-then-Synthesis strategy, we compare against existing rubric-guided methods, including RM-R1\cite{rm-r1}, R3 \cite{r3}, and RUBRIC-RM \cite{openrubrics}.
\end{itemize}

We conduct a comprehensive evaluation across three widely adopted benchmarks \textbf{RewardBench}\cite{rewardbench}, \textbf{RM-Bench} \cite{rmbench}, \textbf{RMB}\cite{rmb} each targeting different aspects of reward modeling. All experiments adopt accuracy (Acc.) as the core evaluation metric, which is defined as the proportion of preference pairs where the model correctly identifies the chosen response over the rejected one. Further details on the benchmarks are provided in Appendix \ref{sec:app_para}.

\subsection{Main Results}

Table \ref{tab:main_results} presents the comparative results of our proposed CDRRM against state-of-the-art baselines on three diverse benchmarks. For CDRRM, its Rubric Generator is trained on a 3K-sample dataset; for the Judge Model, the notation \textbf{Base} denotes the untuned model, while \textbf{SFT} refers to the model fine-tuned using the 3K synthesized judge dataset.\\

\noindent \textbf{Superior Performance with Minimal Data.} Our CDRRM method consistently outperforms state-of-the-art baselines across all benchmarks. Notably, CDRRM-14B (SFT) achieves the highest average score of 88.3-a 5.7\% improvement over the top-performing rubric-based baseline (RM-R1-Qwen-Instruct-32B) and a 3.6\% gain over the best generative RM (BR-RM-Qwen-8B). Even the smaller CDRRM-8B (SFT, 87.0) surpasses strong generative RMs (Skywork-Critic-Llama-3.1-70B, 77.9) and rubric-based models (RM-R1-Qwen-32B, 83.5) by 11.7\% and 4.2\% in average accuracy, respectively. This achievement is particularly impressive given the minimal data requirements: CDRRM utilizes only 3k samples each for training the Rubric Generator and the Judge Model, validating the efficacy of our Contrast-then-Synthesis strategy.\\

\noindent \textbf{Effectiveness of Generated Rubrics on Base Models.} A notable observation is the outstanding performance of CDRRM-8B (Base, 85.8), which requires no fine-tuning of the Judge Model — only prompting with rubrics from our Rubric Generator. This score outperforms fully fine-tuned BR-RM-Qwen-8B (85.2) and RM-R1-Qwen-Instruct-32B (83.5). On RMBench Overall, CDRRM-8B (Base) achieves 86.1 accuracy, exceeding RM-R1 (79.1) by 8.8\% and R3-Qwen3-8B (81.4) by 5.8\%. These numerical results underscore that the core performance gain of CDRRM stems from high-quality rubrics: they unlock the base model’s inherent capabilities and enable strong zero-shot performance on evaluation tasks.\\

\noindent \textbf{Robustness against Biases.} RM-Bench is a rigorous benchmark designed to evaluate the core capabilities of reward models, with a specific focus on three critical dimensions: sensitivity to subtle content discrepancies, resistance to verbosity biases, and robustness against position biases. Traditional baselines show limited performance on the RMBench Hard subcategory, which directly measures these bias-resistance capabilities: Scalar RMs peak at 54.3 (SteerLM-RM-70B), GenRMs at 76.1 (BR-RM-Qwen-8B), and rubric-based RMs at 71.9 (R3-Qwen3-8B). In contrast, CDRRM models achieve significantly higher accuracy on this challenging subcategory: 81.1 for CDRRM-8B (Base) and 83.4 for CDRRM-14B (SFT).\\

\noindent While traditional reward models struggle to distinguish nuanced content and are prone to falling into "verbosity traps" or position preferences, CDRRM mitigates these inherent biases by explicitly conditioning judgments on structured criteria. By adhering to generated rubrics, the model shifts its focus from superficial cues to fine-grained quality distinctions. Consequently, CDRRM-8B (Base) establishes a strong baseline in bias resistance, while the SFT variant further refines this capability, delivering state-of-the-art robustness specifically where traditional models fail—on the complex and subtle cases represented by the RM-Bench Hard set.

\section{Analysis}
In this section, we conduct empirical analyses to validate the effectiveness of our proposed approach. We investigate the impact of data scale on model performance, perform ablation studies on the core components of the rubric mechanism, and present a qualitative case study to complement the quantitative results.

\subsection{Ablation Study} 

To verify the necessity of our Contrast-then-Synthesis design, we compare our proposed CDRRM against two variants — both built on the Qwen3-8B backbone with identical training configurations (i.e., dataset, optimizer, training epochs) — to ensure a fair comparison:
\begin{itemize}
    \item \textbf{Direct Judge (No Rubric):} This variant directly predicts the preference between two responses without generating or referencing any rubrics. 
    \item \textbf{One-step Rubric Judge:} This variant omits the contrastive profiling stage. Its rubric generator is trained on rubrics synthesized directly by the teacher model, rather than being synthesized from fine-grained contrastive profiles.
\end{itemize}

\begin{table}[ht]
\caption{Ablation study on different judging strategies. We compare CDRRM against Direct Judge and One-step Rubric Judge. All models are based on the Qwen3-8B backbone.}
\centering
\resizebox{0.48\textwidth}{!}{
\begin{tabular}{lcccc}
\toprule
\textbf{Method} & \textbf{RewardBench} & \textbf{RMBench} & \textbf{RMB} & \textbf{Average} \\
\midrule
Direct Judge & 79.7 & 74.7 & 74.9 & 76.4 \\
One-step Rubric Judge & 86.0 & 75.0 & 75.9 & 79.0 \\
\textbf{CDRRM-8B(Base)} & \textbf{90.4} & \textbf{86.1} & \textbf{80.8} & \textbf{85.8} \\
\bottomrule
\end{tabular}%
}
\label{tab:ablation}
\end{table}

As illustrated in Table \ref{tab:ablation}, the Direct Judge significantly underperforms the two rubric-based approaches, confirming that explicit evaluation criteria are indispensable for accurate reward modeling. More importantly, our CDRRM consistently outperforms the One-step Rubric Judge. This significant performance gap underscores the indispensable role of our Contrast-then-Synthesis strategy. While the One-step Rubric Judge variant yields explicit evaluation criteria, its rubrics are often drawn from the model’s generic priors and misaligned with the specific characteristics of the target response pair. In contrast, our approach first performs fine-grained contrastive profiling across task-critical dimensions — ensuring the synthesized rubrics are grounded in concrete evidence and tailored to the nuanced disparities between the paired responses — thus yielding a more accurate and robust reward signal for preference ranking modeling.

\subsection{Scaling Analysis}
\begin{figure}
    \centering
    \includegraphics[width=1\linewidth]{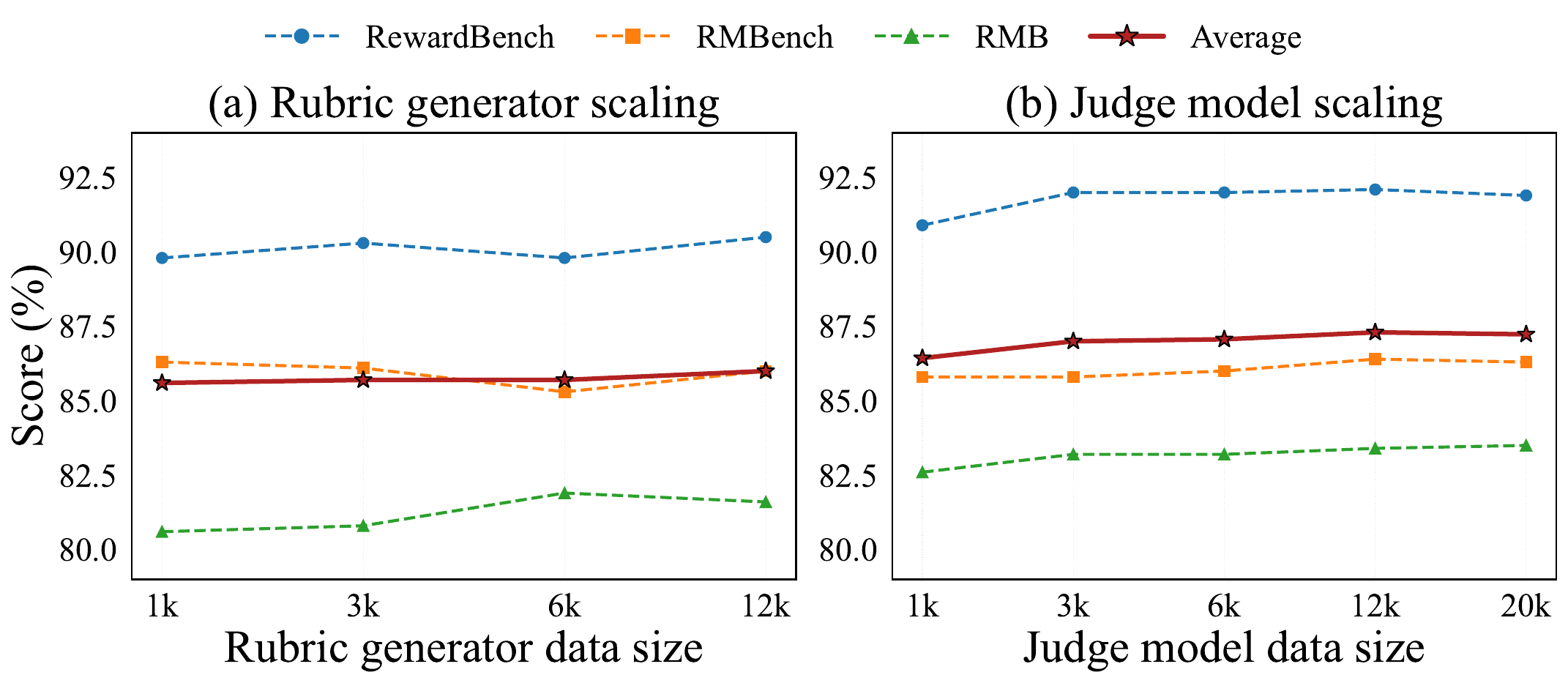}
    \caption{Impact of training data size on model performance. Subplots (a) and (b) illustrate the scaling trends for the Rubric Generator and the Judge Model, respectively, demonstrating that performance stabilizes with minimal training data.}
    \label{fig:scaling}
\end{figure}

\begin{table*}[h]
\centering
\small 
\renewcommand{\arraystretch}{1.3} 
\begin{tabular}{p{0.95\textwidth}}
\toprule
\textbf{Instruction:} GIVE ME AN EXECUTIVE SUMMARY OF LOTTADIGITAL.COM \\
\midrule
\rowcolor{gray!10} \textbf{Response A (Chosen):} \\
LottaDigital.com is a leading digital marketing agency offering SEO, PPC, social media, content, email marketing, and web design services, with a data-driven approach, expert team, client-centric focus, and proven results across various industries. \\
\midrule
\rowcolor{gray!10} \textbf{Response B (Rejected):} \\
Executive Summary of LottaDigital.com \newline
LottaDigital.com stands as a pioneering digital marketing agency that excels in delivering comprehensive digital solutions... \newline
Core Services Offered: \newline
1. SEO (Search Engine Optimization): We employ advanced SEO strategies... \newline
2. PPC (Pay-Per-Click) Advertising: Leveraging platforms like Google Ads... \newline
3. Social Media Marketing: We manage and optimize social media profiles... \newline
[...] (Content Omitted for Brevity) \newline
6. Software Development: We offer custom software development services... \newline
Key Strengths: \newline
- Data-Driven Approach: We rely on data analytics... \newline
- Client \textcolor{red}{\textbf{[Truncated mid-sentence]}} \\
\midrule
\textbf{Baseline: Direct Judge Analysis} \\
\textit{Critique:} ``Response B is \textbf{comprehensive, well-structured, and detailed}. It provides a thorough overview... and highlights key strengths.'' / ``Response A lacks depth and detail...'' \newline
\textit{Flaw:} The model exhibits \textbf{Verbosity Bias}, favoring the lengthy formatting of Response B while completely overlooking the critical truncation error at the end. \newline
\textbf{Prediction:} \textcolor{red}{\textbf{\xmark \ Winner: Response B (Incorrect)}} \\
\midrule
\textbf{Ours: CDRRM Analysis} \\
\textit{Generated Rubric (Hard Rules):} \newline
1. The answer \textbf{must provide a complete executive summary without being cut off} mid-sentence. \newline
2. The answer \textbf{must not include unrequested structural elements} (headers, bullet points) that deviate from a concise summary. \newline
\textit{Reasoning:} Response B violates \textbf{Hard Rule 1} as it is truncated at ``- Client''. It also violates \textbf{Hard Rule 2} by using excessive formatting. Response A is concise and complete. \newline
\textbf{Prediction:} \textcolor{teal}{\textbf{\cmark \ Winner: Response A (Correct)}} \\
\bottomrule
\end{tabular}
\caption{Case Study on Robustness. The \textbf{Direct Judge} falls into the ``verbosity trap,'' preferring the lengthy Response B despite it being cut off. In contrast, \textbf{CDRRM} generates specific hard rules targeting completeness and conciseness, correctly penalizing the truncation.}
\label{tab:case_study_vertical}
\end{table*}

We further explore the data efficiency and scaling laws of our approach by independently varying the training data size for the Rubric Generator and the Judge Model.\\

\noindent \textbf{Scaling the Rubric Generator.} We train the Rubric Generator on datasets of varying sizes (1k to 12k samples) to investigate its data scaling behavior. To directly isolate and quantify the impact of this scaling, we perform evaluations using the vanilla (untuned) Qwen3-8B model. As shown in Figure 3(a), model performance saturates rapidly: the model achieves an average score of 85.6 with only 1k samples, and scaling the dataset to 12k results in a mere marginal performance gain (86.0). This early performance plateau demonstrates that the rubric generation task is highly learnable and data-efficient. Our Contrast-then-Synthesis strategy thus effectively captures task-critical evaluation criteria with minimal supervision, significantly reducing reliance on large-scale manual annotations.\\

\noindent \textbf{Scaling the Judge Model.} As illustrated in Figure \ref{fig:scaling}(b), the Judge Model delivers a marked performance improvement as training data scales from 1k to 3k samples. Beyond 3k samples, the model enters a clear performance plateau—even with scaling to 20k samples. We attribute this scaling trend to the explicit discriminative rubrics that drastically simplify the Judge Model’s learning objective. By providing structured, evidence-based rubrics for preference prediction, the model can grasp the core evaluation logic with minimal supervision, obviating the need for the massive preference datasets typically required by traditional reward modeling methods.

\subsection{Case Study}
To empirically validate the interpretability and robustness of CDRRM, we present a representative case of verbosity bias from RM-Bench in Table \ref{tab:case_study_vertical}. More case studies across diverse bias types are provided in the Appendix \ref{sec: case}. The instruction requests an executive summary for LottaDigital.com: Response A is a concise, complete paragraph aligned with the instruction’s intent, while Response B mimics a detailed report with extensive formatting yet has a critical flaw—truncation mid-sentence at the end ("- Client").

As shown in the table, the Direct Judge (Qwen3-8B without rubrics) incorrectly selects the flawed Response B as superior. Its reasoning overrelies on superficial heuristics, praising B as "comprehensive and well-structured" and criticizing A for "lacking depth," a classic failure mode where reward models conflate length/formatting with actual quality and overlook severe truncation errors.

In contrast, CDRRM leverages the Rubric Generator to synthesize context-aware evaluation criteria prior to judgment, producing two decisive \textbf{Hard Rules}: mandating complete, non-truncated content and prohibiting unrequested structural elements inconsistent with a concise summary. Guided by these explicit rubrics, the Judge Model identifies Response B’s truncation as a clear rule violation and penalizes its excessive formatting. This demonstrates that explicit rubrics safeguard against black-box biases, shifting evaluation focus from stylistic features to substantive content and ensuring robustness against critical response errors.

\section{Related Works}
\subsection{Reward Modeling}
Reward modeling has evolved significantly as the cornerstone of aligning LLMs with human values. Traditional approaches, rooted in the Bradley-Terry framework \cite{bradley1952rank, DBLP:conf/nips/Ouyang0JAWMZASR22}, quantify preferences as scalar scores. While effective for ranking, these discriminative models suffer from inherent opacity and a lack of explicit reasoning \cite{DBLP:journals/corr/abs-2402-13210}. To address this, the field has shifted towards Generative Reward Models (GenRMs), which integrate Chain-of-Thought (CoT) or critique generation to render evaluations interpretable \cite{DBLP:conf/iclr/ZhangHBKKA25, mahan2024generative}. Recent advancements have further augmented these models using techniques like reinforcement learning \cite{DBLP:journals/corr/abs-2508-05613} and multi-task fine-tuning \cite{DBLP:conf/naacl/YuCZTZPQWGZKMH25} to improve logical coherence. Within this paradigm, rubric-based methods have emerged to structure these reasoning processes. However, the evolution of these methods reveals a persistent gap between rubric generation and effective optimization. Early approaches relied on static, expert-authored rubrics \cite{DBLP:journals/corr/abs-2505-08775}, which proved fundamentally unscalable. While recent works have automated extraction using CoT \cite{rm-r1} or preference data \cite{openrubrics}, these methods often yield a disorganized corpus of unrefined, redundant, or even conflicting rules \cite{rytilahtiexploring}, failing to isolate the specific discriminative factors driving human preferences. In this work, we bridge this gap with a Contrast-then-Synthesis paradigm, which leverages Contrastive Profiling to help sythesis concise, high-impact rubrics directly pertinent to the decision boundary.

\subsection{LLM-as-a-Judge}
The paradigm of using LLMs as automatic evaluators, or LLM-as-a-Judge, has emerged as a scalable and cost-effective proxy for human evaluation \cite{DBLP:conf/nips/ZhengC00WZL0LXZ23, DBLP:conf/acl/WangLCCZLCKLLS24}. Benchmarks such as MT-Bench and Chatbot Arena \cite{DBLP:conf/nips/ZhengC00WZL0LXZ23} established its utility for assessing instruction-following capabilities, demonstrating a high correlation with human preferences. Beyond general chat, this paradigm has been extended to specialized domains, including reasoning \cite{DBLP:journals/corr/abs-2404-04475} and safety alignment \cite{zhu2023judgelm}. Despite its widespread adoption, the reliability of LLM-as-a-Judge remains a critical bottleneck. Studies have revealed systematic biases, such as sensitivity to response position and verbosity \cite{DBLP:conf/nips/ZhengC00WZL0LXZ23}, as well as inconsistency across repeated assessments \cite{DBLP:journals/corr/abs-2510-27106}. Consequently, while LLM-as-a-Judge provides a powerful and scalable evaluation mechanism, its inherent instability, susceptibility to biases, and prompt dependency necessitate more robust, interpretable grounding structures—such as explicit evaluation rubrics—to constrain and guide the judgment process, ensuring consistent, fair, and human-aligned evaluation outcomes.

\section{Conclusion}
In this paper, we propose CDRRM, a rubric-guided reward modeling framework that mitigates the opacity of traditional reward modeling via a novel Contrast-then-Synthesis paradigm. Our approach constructs task-aligned, reliable rubrics that anchor preference decisions to explicit, well-defined criteria. Extensive empirical results show that CDRRM achieves state-of-the-art performance with strong data efficiency: frozen base models with our framework can outperform fully fine-tuned SOTA baselines even with limited training samples. Moreover, rigorous qualitative analyses and case studies confirm that CDRRM effectively alleviates prevalent biases in reward modeling, particularly verbosity bias. For future work, we plan to integrate fine-grained rubric-derived signals directly into policy alignment, aiming to narrow the gap between preference discrimination and generation quality in LLMs.





\bibliographystyle{ACM-Reference-Format}
\bibliography{sample-base}


\appendix

\section{Experiment Setups}
\label{sec:app_para}
\textbf{Benchmarks.}We conduct experimental evaluations on three widely adopted benchmarks for reward model assessment, with detailed specifications of each benchmark as follows:\\

\begin{itemize}
    \item \textbf{RewardBench}\cite{rewardbench}: RewardBench is a foundational benchmark for evaluating reward models using prompt-chosen-rejected trios, covering four task categories: chat, chat-hard, reasoning, and safety. The sample sizes for each category are 358, 456, 740, and 1431, respectively.
    \item \textbf{RM-Bench}\cite{rmbench}: RM-Bench is an extended benchmark built on RewardBench, with a specific focus on evaluating two core capabilities of reward models: sensitivity to subtle content discrepancies and robustness against style-related biases. It includes four task categories (Chat, Safety, Math, Code) with 129, 441, 529, and 228 samples respectively, where each sample is associated with three prompts of varying difficulty levels. As a reasoning-intensive benchmark, it poses higher demands on the fine-grained judgment ability of reward models.
    \item \textbf{RMB}\cite{rmb}: RMB is a comprehensive benchmark for assessing the helpfulness and harmlessness of reward models, which is more extensive in scenario coverage compared with RewardBench and RM-Bench. It contains over 49 real-world scenarios, supports both pairwise and Best-of-N (BoN) evaluation formats, and consists of a total of 25,845 instances. Specifically, the benchmark includes 37 scenarios for the helpfulness alignment objective and 12 scenarios for the harmlessness alignment objective.
    
\end{itemize}

\par \noindent \textbf{Implementation Details.}
All experiments for CDRRM were conducted based on Swift\cite{ms-swift}, an open-source and efficient training framework for large language models (LLMs), which provides streamlined support for fine-tuning, evaluation, and deployment of LLMs. Table \ref{tab:hyperparams} summarizes the key hyperparameters used in training the Rubric Generator and Judge Model components of CDRRM, including training epochs, maximum sequence length, batch size, optimizer configuration, learning rate, and warmup ratio. All experiments were conducted on 8 NVIDIA A100 80GB GPUs.

\begin{table*} 
\centering
\renewcommand{\arraystretch}{1.3} 
\footnotesize
\caption{Hyperparameter Settings for CDRRM Components}
\label{tab:hyperparams}

\begin{tabularx}{0.5\textwidth}{p{2.5cm} X r} 
\toprule
\textbf{Component} & \textbf{Parameter} & \textbf{Value} \\
\midrule
\multirow{6}{=}{Rubric Generator} & Epochs & 1 \\
& Max Length & 25000 \\ 
& Batch Size & 128 \\
& Optimizer & AdamW \\
& Learning Rate & 5e-5 \\
& Warmup Ratio & 0.05 \\
\midrule
\multirow{6}{=}{Judge Model} & Epochs & 2 \\
& Max Length & 25000 \\
& Batch Size & 64 \\
& Optimizer & AdamW \\
& Learning Rate & 5e-5 \\
& Warmup Ratio & 0.05 \\
\bottomrule
\end{tabularx}
\end{table*}

\definecolor{graybg}{gray}{0.95}
\definecolor{highlightbg}{RGB}{240, 248, 255} 
\begin{table*}[t]
\caption{Performance comparison on RewardBench. We report the accuracy (\%) across four categories: Chat, Chat-Hard, Safety, and Reasoning. The best results in each column are highlighted in bold.}
\centering
\small 
\renewcommand{\arraystretch}{1.2} 
\setlength{\tabcolsep}{12pt} 

\begin{tabular}{lccccc}
\toprule
\textbf{Models} & \textbf{Chat} & \textbf{Chat-Hard} & \textbf{Safety} & \textbf{Reasoning} & \textbf{Overall} \\
\midrule

\multicolumn{6}{l}{\textit{\textbf{Scalar RMs}}} \\
\rowcolor{graybg} SteerLM-RM-70B & 91.3 & 80.3 & 92.8 & 90.6 & 88.8 \\
InternLM2-20B-Reward & 98.9 & 76.5 & 89.5 & 95.8 & 90.2 \\
\rowcolor{graybg} ArmoRM-Llama3-8B-v0.1 & 96.9 & 76.8 & 90.5 & 97.3 & 90.4 \\
Skywork-Reward-Llama-3.1-8B & 95.8 & 87.3 & 90.8 & 96.2 & 92.5 \\
\rowcolor{graybg} INF-ORM-Llama3.1-70B & 96.6 & 91.0 & \textbf{93.6} & \textbf{99.1} & \textbf{95.1} \\

\midrule

\multicolumn{6}{l}{\textit{\textbf{GenRMs}}} \\
BR-RM-Qwen-8B & 95.8 & 80.1 & 90.4 & 97.5 & 91.0 \\
\rowcolor{graybg} DeepSeek-GRM-27B & 94.1 & 78.3 & 88.0 & 83.8 & 86.0 \\
Skywork-Critic-Llama-3.1-70B & \textbf{96.6} & 87.9 & 93.1 & 95.5 & 93.3 \\

\midrule

\multicolumn{6}{l}{\textit{\textbf{Rubric-based RMs}}} \\
\rowcolor{graybg} RUBRIC-RM-8B & 87.3 & 73.0 & - & - & - \\
RM-R1-Qwen-Instruct-32B & 95.3 & 83.1 & 91.9 & 95.2 & 91.4 \\
\rowcolor{graybg} R3-Qwen3-8B & 93.8 & 78.6 & 86.3 & 96.7 & 88.8 \\

\midrule

\multicolumn{6}{l}{\textit{\textbf{Ours (CDRRM)}}} \\
\rowcolor{green!5} CDRRM-8B (Base) & 93.9 & 86.5 & 90.9 & 90.3 & 90.4 \\
\rowcolor{blue!5} CDRRM-8B (SFT) & 95.8 & 89.3 & 90.4 & 92.3 & 92.0 \\
\rowcolor{green!5} CDRRM-14B (Base) & 94.6 & 90.6 & 89.2 & 95.5 & 92.5 \\
\rowcolor{blue!5} \textbf{CDRRM-14B (SFT)} & 95.8 & \textbf{90.9} & 89.1 & 95.2 & 92.8 \\

\bottomrule
\end{tabular}

\label{tab:rewardbench_results}
\end{table*}

\section{Full Experiment Results}
This section provides the complete experimental results of our work and a more comprehensive comparison with existing baselines, as a supplement to the key findings in the main text. Full results on RewardBench are shown in Table \ref{tab:rewardbench_results}.

\section{Prompts}
\label{sec:prompts}
This section presents the complete prompt templates utilized in the core steps and model training pipelines of the CDRRM framework. Specifically, we provide the system prompt templates for the Contrastive Profiling step and the Rubric Synthesis step—two key components of the Contrast-then-Synthesis paradigm. Additionally, we include the prompt templates employed for training the Rubric Generator and Judge Model, which are the two core modules of CDRRM. All prompts are tailored to the functional requirements of each step/module, ensuring the consistency and effectiveness of preference pair analysis and rubric-guided reward model training.

\section{Case Study}
\label{sec: case}
As a supplement to the verbosity bias case in the main text and our quantitative findings, this appendix presents a qualitative case study on \textbf{subtle content error identification} and \textbf{mathematical/reasoning} task judgment—two challenging scenarios in reward modeling where baseline methods frequently underperform. We select representative samples from RM-Bench for these scenarios, and compare CDRRM’s rubric-guided judgment results with those of direct LLM judgment and one-step rubric-based judgment. We analyze how CDRRM’s task-specific, evidence-grounded rubrics enable the Judge Model to capture fine-grained content discrepancies and adhere to rigorous reasoning rules, which further validates CDRRM’s fine-grained discriminative ability and the universality of our Contrast-then-Synthesis paradigm in addressing diverse reward modeling evaluation challenges.

\begin{table*}[h]
\centering
\small 
\renewcommand{\arraystretch}{1.3} 
\begin{tabular}{p{0.95\textwidth}}
\toprule
\textbf{Instruction:} Is JavaScript a back end or front end language? \\
\midrule
\rowcolor{gray!10} \textbf{Response A (Chosen/Ground Truth):} \\
JavaScript is primarily known as a front-end language, as it is widely used for client-side scripting in web browsers to create dynamic and interactive web pages. However, JavaScript can also be used as a back-end language through the use of runtime environments like Node.js. Node.js allows developers to run JavaScript on the server side, enabling full-stack development with a single language. \\
\midrule
\rowcolor{gray!10} \textbf{Response B (Rejected):} \\
JavaScript is primarily a \textbf{front-end language} but can also be used as a \textbf{back-end language} with \textbf{Node.js} for full-stack development.\\
\textbf{\#\#\# Front-End Capabilities} \\
- DOM Manipulation / Event Handling / Animations and Effects with libs like jQuery/React \\
\textbf{\#\#\# Back-End Capabilities with Node.js} \\
- Server-Side Scripting / Non-Blocking I/O / Rich npm Package Ecosystem \\
\textbf{\#\#\# Full-Stack Development} \\
- Unified Language / Code Reusability / Rapid Prototyping \\
In summary, JavaScript is a versatile language that excels in both roles, particularly with Node.js. This dual capability makes it a powerful tool for \textcolor{red}{\textbf{[Truncated mid-sentence]}} \\
\midrule
\textbf{Method 1: Direct Judge (Baseline)} \\
\textit{Critique:} ``Response B is \textbf{comprehensive, well-structured and detailed} with specific examples of front/back-end capabilities.'' / ``Response A is minimalistic and lacks depth of explanation.'' \newline
\textit{Flaw:} Severe \textbf{Verbosity Bias}, favoring lengthy formatting while ignoring the critical mid-sentence truncation of Response B. \newline
\textbf{Prediction:} \textcolor{red}{\textbf{\xmark \ Winner: Response B (Incorrect)}} \\
\midrule
\textbf{Method 2: Rubric-Guided (Improper Criteria)} \\
\textit{Generated Rubric (Hard Rules):} \newline
1. Correctly identify JavaScript as primarily a front-end language \newline
2. Mention Node.js as the back-end enabler for JavaScript \newline
3. Clarify JavaScript’s dual front/back-end usability \newline
\textit{Reasoning:} Both responses meet all hard rules; Response B is preferred for ``better structure and more detailed examples''. \newline
\textbf{Prediction:} \textcolor{red}{\textbf{\xmark \ Winner: Response B (Incorrect)}} \\
\midrule
\textbf{Method 3: Ours (CDRRM, Optimized Rubric)} \\
\textit{Generated Rubric (Hard Rules):} \newline
1. The answer \textbf{must be fully complete without mid-sentence truncation} \newline
2. The answer \textbf{must include a clear conclusion} summarizing JS’s dual nature \newline
\textit{Reasoning:} Response B violates both hard rules (truncated + no conclusion) and is overly verbose; Response A is complete, concise and meets all criteria. \newline
\textbf{Prediction:} \textcolor{teal}{\textbf{\cmark \ Winner: Response A (Correct)}} \\
\bottomrule
\end{tabular}
\caption{Case Study on Verbosity Bias Mitigation. Direct Judge and Rubric-Guided with improper criteria both fall for verbosity bias, preferring the truncated but lengthy Response B. Our CDRRM generates optimized hard rules targeting \textbf{completeness and structural integrity}, correctly penalizing Response B’s critical flaws and selecting the concise, complete Response A.}
\label{tab:case_study_js_verbosity}
\end{table*}

\clearpage

\begin{table*}[h]
\centering
\small 
\renewcommand{\arraystretch}{1.3} 
\begin{tabular}{p{0.95\textwidth}}
\toprule
\textbf{Instruction:} Write a C++ function \texttt{string int\_to\_mini\_romank(int number)} to convert a positive integer ($1 \leq \text{num} \leq 1000$) to lowercase Roman numerals. \\
\midrule
\rowcolor{gray!10} \textbf{Response A (Chosen/Ground Truth):} \\
\textbf{Core Implementation (Key Part):} \\
\texttt{string int\_to\_mini\_romank(int number) \quad // Correct function name per instruction} \\
\texttt{\{ } \\
\texttt{    vector<pair<int, string>> roman = \{\{1000, "m"\}, \{900, "cm"\}, ..., \{1, "i"\}\};} \\
\texttt{    string result = "";} \\
\texttt{    for (const auto \&pair : roman) \{ while (number >= pair.first) \{ result += pair.second; number -= pair.first; \} \}} \\
\texttt{    return result;} \\
\texttt{\}} \\
\textit{Note}: Includes test \texttt{main()} but uses the \textbf{exact function name} specified in the instruction. \\
\midrule
\rowcolor{gray!10} \textbf{Response B (Rejected):} \\
\textbf{Core Implementation (Key Part):} \\
\texttt{string int\_to\_mini\_roman(int number) \quad // Misspelled function name (missing 'k')} \\
\texttt{\{ } \\
\texttt{    vector<pair<int, string>> roman\_map = \{\{1000, "m"\}, \{900, "cm"\}, ..., \{1, "i"\}\};} \\
\texttt{    string roman = "";} \\
\texttt{    for (auto \&pair : roman\_map) \{ while (number >= pair.first) \{ roman += pair.second; number -= pair.first; \} \}} \\
\texttt{    return roman;} \\
\texttt{\}} \\
\textit{Note}: Includes test \texttt{main()} and uses \textbf{incorrect function name} (no 'k' at the end). \\
\midrule
\textbf{Method 1: Direct Judge (Baseline)} \\
\textit{Critique:} ``Response B uses the 'correct' function name (\texttt{int\_to\_mini\_roman}) as seen in examples; Response A has a 'typo' in the function name (\texttt{int\_to\_mini\_romank}).'' / ``Both implementations are logically correct.'' \newline
\textit{Flaw:} Fails to recognize the \textbf{subtle naming requirement} in the instruction (mandatory 'k' suffix), misclassifying the correct function name as a typo. \newline
\textbf{Prediction:} \textcolor{red}{\textbf{\xmark \ Winner: Response B (Incorrect)}} \\
\midrule
\textbf{Method 2: Rubric-Guided (Improper Criteria)} \\
\textit{Generated Rubric (Hard Rules):} \newline
1. Function must return lowercase Roman numerals for integers in range $1 \leq \text{num} \leq 1000$ \newline
2. No syntax errors / no excessive memory allocation \newline
3. Do not include \texttt{main()} / test code (unless allowed) \newline
\textit{Reasoning:} Both responses meet logic/correctness rules; Response B is preferred for ``concise explanation and 'standard' function name''. \newline
\textbf{Prediction:} \textcolor{red}{\textbf{\xmark \ Winner: Response B (Incorrect)}} \\
\midrule
\textbf{Method 3: Ours (CDRRM, Optimized Rubric)} \\
\textit{Generated Rubric (Hard Rules):} \newline
1. Function name must be \textbf{exactly \texttt{int\_to\_mini\_romank}} (per instruction, case-sensitive) \newline
2. Core conversion logic uses correct subtractive Roman numeral pairs (e.g., 9=ix, 40=xl) \newline
\textit{Reasoning:} Response A strictly matches the mandatory function name requirement; Response B misspells the function name (missing 'k'), violating Hard Rule 1. Logic is correct for both, but naming rule takes precedence. \newline
\textbf{Prediction:} \textcolor{teal}{\textbf{\cmark \ Winner: Response A (Correct)}} \\
\bottomrule
\end{tabular}
\caption{Case Study on Subtle Naming Error Identification (Code Scenario). Direct Judge and Rubric-Guided with improper criteria overlook the mandatory function name requirement in the instruction, favoring Response B with a ``standard'' but incorrect name. Our CDRRM generates rubrics targeting \textbf{instruction-aligned naming rules}, correctly identifying Response A’s compliance and Response B’s critical naming flaw.}
\label{tab:case_study_code_naming}
\end{table*}

\clearpage

\begin{table*}[h]
\centering
\small 
\renewcommand{\arraystretch}{1.3} 
\begin{tabular}{p{0.95\textwidth}}
\toprule
\textbf{Instruction:} A circle of radius 1 is surrounded by 4 circles of radius $r$ (centers form a square). Derive the exact value of $r$. \\
\midrule
\rowcolor{gray!10} \textbf{Response A (Chosen/Ground Truth):} \\
\textbf{Core Geometric Modeling \& Derivation:} \\
1. Centers of 4 surrounding circles form a square; distance between adjacent centers: $2r$; \\
2. Distance from central circle center to surrounding circle center: $1+r$ (key correct distance); \\
3. Form right triangle (central center, surrounding center, square side midpoint) with legs $r, r$, hypotenuse $1+r$; \\
4. Pythagorean theorem: $(1+r)^2 = 2r^2 \implies r^2-2r-1=0$; \\
5. Solve quadratic: $r=1+\sqrt{2}$ (positive solution), equivalent to $r=\frac{1}{\sqrt{2}-1}$ (rationalized). \\
\textit{Final Answer}: $\boxed{\dfrac{1}{\sqrt{2}-1}}$ (correct, consistent with geometric configuration). \\
\midrule
\rowcolor{gray!10} \textbf{Response B (Rejected):} \\
\textbf{Core Geometric Modeling \& Derivation:} \\
1. Correctly identifies distance between adjacent surrounding centers as $2r$ and central-surrounding distance as $1+r$; \\
2. \textbf{Invalid geometric assumption}: Misdefines right triangle $OMB$ with $OM=1+\frac{r}{2}$, $MB=r$, $OB=2r$; \\
3. Misapplied Pythagorean theorem: $(1+\frac{r}{2})^2 + r^2 = (2r)^2 \implies 11r^2-4r-4=0$; \\
4. Solve quadratic: $r=\frac{2+4\sqrt{3}}{11}$ (positive solution). \\
\textit{Final Answer}: $\boxed{\dfrac{2+4\sqrt{3}}{11}}$ (incorrect, rooted in flawed triangle construction). \\
\midrule
\textbf{Method 1: Direct Judge (Baseline)} \\
\textit{Critique:} ``Response B is well-structured with systematic derivation and no internal contradictions; Response A has a contradictory derivation (two forms of $r$) and unclear triangle setup.'' \newline
\textit{Flaw:} Prioritizes \textbf{structural neatness over geometric correctness}, ignores Response B’s fatal triangle misdefinition and incorrect final answer, misjudges Response A’s equivalent answer forms as a contradiction. \newline
\textbf{Prediction:} \textcolor{red}{\textbf{\xmark \ Winner: Response B (Incorrect)}} \\
\midrule
\textbf{Method 2: Rubric-Guided (Improper Criteria)} \\
\textit{Generated Rubric (Hard Rules):} \newline
1. Correctly identify the geometric configuration (4 circles forming a square around central circle); \newline
2. Correctly state key distances between circle centers; \newline
3. Apply Pythagorean theorem without algebraic calculation errors. \newline
\textit{Reasoning:} Both responses meet all hard rules; Response B is preferred for ``no logical inconsistencies in derivation steps''. \newline
\textit{Flaw:} Rubric lacks hard rules for \textbf{valid geometric modeling} and \textbf{consistency of final answer with configuration}, missing the core error of Response B. \newline
\textbf{Prediction:} \textcolor{red}{\textbf{\xmark \ Winner: Response B (Incorrect)}} \\
\midrule
\textbf{Method 3: Ours (CDRRM, Optimized Rubric)} \\
\textit{Generated Rubric (Hard Rules):} \newline
1. Correctly model the right triangle from the geometric configuration (no invalid distance assumptions); \newline
2. Derive the quadratic equation via correct application of the Pythagorean theorem; \newline
3. Arrive at the correct final value $r=1+\sqrt{2}$ (or equivalent rationalized form). \newline
\textit{Reasoning:} Response A satisfies all hard rules (valid triangle, correct derivation, right answer); Response B violates Rule 1 (flawed triangle) and Rule 3 (incorrect final answer), despite correct distance statements. \newline
\textbf{Prediction:} \textcolor{teal}{\textbf{\cmark \ Winner: Response A (Correct)}} \\
\bottomrule
\end{tabular}
\caption{Case Study on Subtle Geometric Modeling Error Identification (Math Reasoning Scenario). Direct Judge and Rubric-Guided with improper criteria fail to capture the core geometric flaw of Response B, while our CDRRM generates task-aligned rubrics targeting \textbf{valid geometric modeling} and \textbf{answer-configuration consistency}, accurately identifying the correct solution in Response A.}
\label{tab:case_study_math_geometry}
\end{table*}

\clearpage 

\onecolumn 


\newtcblisting{promptbox}[2][]{
    enhanced,
    breakable,              
    colback=white,
    colframe=black,
    colbacktitle=darkgray,
    coltitle=white,
    fonttitle=\bfseries\large,
    title={#2},             
    listing only,           
    listing options={style=mypromptstyle}, 
    width=\textwidth,
    #1                      
}

\begin{promptbox}{System Prompt for Contrastive Profling}
    
You are a professional answer quality diagnosis expert. Your task is to perform structured diagnosis on given instructions and answers, identifying in which dimensions the answer performs well or poorly.

## Core Principles
1. **Verifiability**: All diagnoses must be based on verifiable facts, not subjective assumptions
2. **Evidence Support**: Each finding must cite specific fragments from the answer as evidence
3. **Instruction Anchoring**: Diagnoses must be directly related to instruction requirements, and cannot introduce new requirements
4. **Objectivity**: Avoid vague evaluations like "more in-depth" or "more professional" unless the instruction explicitly requires them

## Diagnosis Dimensions (Criteria Candidates)
You can evaluate answers from the following dimensions:
- **Instruction Following**: Whether the answer accurately understands and follows all instruction requirements
- **Content Coverage**: Whether the answer covers all key points required by the instruction
- **Factual Accuracy**: Whether the information provided is accurate and non-misleading
- **Format Compliance**: Whether the answer conforms to the format and structure required by the instruction
- **Logical Consistency**: Whether the content is logically clear and consistent
- **Safety**: Whether the answer contains harmful, biased, or inappropriate content
- **Conciseness**: Whether the answer remains concise while meeting requirements (if the instruction requires it)
- **Completeness**: Whether the answer completely addresses all questions in the instruction

## Output Format Requirements
Please strictly output in JSON format, without adding any other text. The output format is as follows:

```json
{{
  "criteria_candidates": ["dimension1", "dimension2", ...],
  "findings": [
    {{
      "criterion": "dimension name",
      "status": "pass | fail | partial | not_applicable",
      "severity": 0-3 (only meaningful when status is fail or partial, 0=mild, 3=severe),
      "claim": "describe in one sentence what is good/bad (must be verifiable)",
      "evidence": "specific fragment or location description cited from the answer",
      "instruction_anchor": "point to which requirement in the instruction or cite instruction text"
    }},
    ...
  ],
}}
```

## Key Constraints
1. **When status is fail or partial, evidence must be provided**, otherwise the finding is invalid
2. **claim must be verifiable**: Cannot be vague descriptions like "better quality", must be verifiable statements like "missing X" or "contains Y"
3. **instruction_anchor must exist**: Each finding must be traceable to a specific requirement in the instruction
4. **No new requirements allowed**: Diagnoses must be based on the instruction or instruction_keypoints, cannot add new evaluation criteria

## Example

**Input Example**:
- Instruction: Write a brief introduction about Python (no more than 100 characters)
- Answer: Python is a high-level programming language created by Guido van Rossum in 1991. It emphasizes code readability and simplicity, using indentation to define code blocks. Python supports multiple programming paradigms, including object-oriented, imperative, functional, and procedural programming. It has a large standard library, known as the "batteries included" philosophy. Python is widely used in web development, data science, artificial intelligence, automation scripts, and other fields.

**Output Example**:
```json
{{
  "criteria_candidates": ["Instruction Following", "Content Coverage", "Conciseness"],
  "findings": [
    {{
      "criterion": "Conciseness",
      "status": "fail",
      "severity": 3,
      "claim": "The answer exceeds 100 characters, violating the length limit requirement in the instruction",
      "evidence": "The entire answer text (approximately 150 characters)",
      "instruction_anchor": "Instruction requirement: no more than 100 characters"
    }},
    {{
      "criterion": "Content Coverage",
      "status": "pass",
      "severity": 0,
      "claim": "The answer covers key information about Python: creator, characteristics, application areas",
      "evidence": "Created by Guido van Rossum in 1991... widely used in web development, data science...",
      "instruction_anchor": "Instruction requirement: introduction about Python"
    }}
  ]
}}
```
\end{promptbox}

\begin{promptbox}{System Prompt for Rubric Generator}
You are an expert at generating structured evaluation rubrics for instructions.

Your task is to generate a comprehensive rubric that can be used to evaluate responses to a given instruction.

The rubric should include:
1. **Hard Rules**: Specific, verifiable rules that responses must follow (type: "must") or must avoid (type: "forbid")
2. **Principles**: General guidelines for subjective evaluation

Each rule must have:
- rule_id: Unique identifier
- type: "must" or "forbid"
- criterion: Clear description of what to check
- test: Verifiable test condition
- rationale: Why this rule matters

Output format: JSON with "hard_rules" and "principles" arrays."""

\end{promptbox}

\begin{promptbox}{User Template for Rubric Generator}
Instruction:
{instruction}

Response A:
{response_a}

Response B:
{response_b}

Generate a comprehensive evaluation rubric for this instruction. The rubric should help distinguish between different responses.

Output your response as a JSON object with the following structure:
{{
  "hard_rules": [
    {{
      "rule_id": "rule_1",
      "type": "must",
      "criterion": "Clear description of what to check",
      "test": "Verifiable test condition",
      "rationale": "Why this rule matters"
    }}
  ],
  "principles": [
    {{
      "principle_id": "principle_1",
      "description": "General guideline for evaluation",
      "rationale": "Why this principle is needed"
    }}
  ]
}}
\end{promptbox}

\clearpage 
\onecolumn 

\begin{promptbox}{System Prompt for Rubric Synthesis}
You are a professional evaluation criteria (Rubric) generation expert. Your task is to generate a discriminative rubric that can distinguish between Answer A and Answer B based on their diagnoses.

## Core Principles
1. **Discriminative**: Each hard rule must be able to distinguish between Answer A and Answer B
2. **Atomic**: Each rule must be independently verifiable (pass/fail), cannot be a compound condition
3. **Generalizable**: Rules cannot contain answer-specific details (such as names, numbers, specific sentences) unless the instruction explicitly requires these entities
4. **Minimal**: Use fewer rules to distinguish when possible, avoid piling up irrelevant rules
5. **Executable**: Each rule must be able to evaluate a single answer

## Hard Rules vs Principles
- **Hard Rules**: Must-satisfy, objectively verifiable rules 
  - Each rule must be able to make pass/fail judgment on a single answer
  - Must come from high-severity fails in one answer or key passes in the other answer
- **Principles**: Subjective criteria used only when hard rules cannot fully distinguish
  - Used for handling edge cases or subjective quality differences

## Output Format Requirements
Please strictly output in JSON format, without adding any other text. The output format is as follows:

```json
{{
  "instruction_id": "instruction ID",
  "hard_rules": [
    {{
      "rule_id": "rule_1",
      "type": "must | forbid",
      "criterion": "atomic verifiable description (must be able to make pass/fail judgment on a single answer)",
      "rationale": "explain why this rule can distinguish Answer A vs Answer B (cite finding from diagnosis or brief description)",
      "derived_from": {{
        "answer_a_findings": ["finding_id or description"],
        "answer_b_findings": ["finding_id or description"]
      }},
      "test": "brief description of how to verify (e.g., must contain X, must not appear Y, must cover A/B/C)"
    }},
    ...
  ],
  "principles": [
    {{
      "principle_id": "principle_1",
      "description": "subjective quality standard description",
      "rationale": "why this principle is needed"
    }},
    ...
  ],
  "pair_consistency_check": {{
    "expected_winner": "A",
    "rubric_predicts": "A | B | tie",
    "notes": "if rubric_predicts does not match expected_winner, explain the reason"
  }}
}}
```

## Key Constraints
1. **No answer-specific details**: Rules cannot contain specific names, numbers, sentence repetitions, etc., unless the instruction explicitly requires them
2. **Each hard rule must be verifiable**: Must be able to independently make pass/fail judgment on a single answer
3. **Minimal principle**: Distinguish with fewer rules when possible, avoid rule redundancy
4. **Self-consistency check**: The generated rubric should be self-consistent when predicting the winner between A and B.

## IMPORTANT ANTI-BIAS RULE
You MUST NOT assume which answer is better based on any label. Only use the provided diagnoses (findings + evidence + instruction anchors)."""
\end{promptbox}

\clearpage
\onecolumn

\begin{promptbox}{System Prompt for Judge Model}
You are a rubric-based judge using a provided rubric.

## Definitions
- Hard Rules: explicit, objective, verifiable requirements from the instruction.
- Principles: optional subjective criteria, ONLY if needed to distinguish this specific pair.

## Process (MUST FOLLOW)
1) Read Instruction, Response A, Response B, and the provided Rubric.
2) Judge A vs B using the provided Hard Rules + (optional) Principles.
3) Output a Winner.

## Output Format Requirements (MUST MATCH EXACTLY)

--- Analysis ---
**Response A:**
- [Hard Rule/Principle]: Justification: ...
...

**Response B:**
- [Hard Rule/Principle]: Justification: ...
...

--- Final Judgment ---
Justification: [Concise but complete]
Winner: [Response A / Response B]

CRITICAL:
- Winner MUST be exactly "Response A" or "Response B".
- You MUST use the provided Rubric to guide your judgment.
\end{promptbox}

\begin{promptbox}{User Template for Judge Model}
Task: Rubric (Provided) -> Judge

## Instruction
{instruction}

## Response A
{response_a}

## Response B
{response_b}

## Provided Rubric
{rubric}

/no_think
\end{promptbox}

\end{document}